\documentclass[letterpaper, 10 pt, conference]{ieeeconf}  % Comment this line out
\IEEEoverridecommandlockouts                              % This command is only
\usepackage{amsmath,amssymb,amsfonts,booktabs,multirow,bm,balance,bbm,siunitx,hyperref,graphicx,algorithm}
\usepackage[noend]{algpseudocode}
\usepackage[caption=false, font=footnotesize]{subfig}
\usepackage[usenames,dvipsnames]{color}

\newcommand{\bs}{\boldsymbol}
\newcommand{\roundB}[1]{\left( #1 \right)} %enclose in round brackets
\newcommand{\skewSymm}[1]{\bs S\!\roundB{#1}} %enclose in | . |

\newcommand{\jerk}{\bs{\dddot{x}}}
\newcommand{\acc}{\bs{\ddot{x}}}
\newcommand{\vel}{\bs{\dot{x}}}
\newcommand{\pos}{\bs x}

\newcommand{\gravity}{\bs g}

\newcommand{\thrust}{f}

\newcommand{\zDir}{\bs e_3}

\newcommand{\rotation}{\bs R}
\newcommand{\rotationDot}{\bs{\dot{R}}}
\newcommand{\angVel}{\bs \omega}

\newcommand{\finalTime}{T}
\newcommand{\splitTime}{t_{\mathrm{split}}}
\newcommand{\sectionStartTime}{t_s}
\newcommand{\sectionEndTime}{t_f}
\newcommand{\minTimeSection}{t_{\mathrm{min}}}
\newcommand{\planePos}{\bs p}
\newcommand{\dir}{\bs n}
\newcommand{\dist}{d}
\newcommand{\extSet}{\mathcal{T}_{\mathrm{crit}}}
\newcommand{\extSetHigh}{\mathcal{T}_{\mathrm{crit}}^{(\uparrow)}}
\newcommand{\extSetLow}{\mathcal{T}_{\mathrm{crit}}^{(\downarrow)}}
\newcommand{\distDot}{\dot{d}}
\newcommand{\obs}{\mathcal{O}}

\newcommand{\quadRadius}{r_{q}}
\newcommand{\obsTraj}{\bs x_\mathcal{O}}
\newcommand{\diffTraj}{\bs{\tilde{x}}}
\newcommand{\projPos}{\bs{x}_p}
\newcommand{\projVel}{\bs{\dot{x}}_p}

\newcommand{\finalPos}{\bs{p}_f}

\title{\LARGE \bf
Rapid Collision Detection for Multicopter Trajectories
}

\author{Nathan Bucki and Mark W. Mueller% <-this % stops a space
\thanks{The authors are with the High Performance Robotics Lab, University of California, Berkeley, CA 94703, USA.
{\tt\small \{nathan\_bucki, mwm\}@berkeley.edu} }
}

\begin{document}

\maketitle
\thispagestyle{empty}
\pagestyle{empty}

%%%%%%%%%%%%%%%%%%%%%%%%%%%%%%%%%%%%%%%%%%%%%%%%%%%%%%%%%%%%%%%%%%%%%%%%%%%%%%%%
\begin{abstract}
We present a continuous-time collision detection algorithm for quickly detecting whether certain polynomial trajectories in time intersect with convex obstacles.
The algorithm is used in conjunction with an existing multicopter trajectory generation method to achieve rapid, obstacle-aware motion planning in environments with both static convex obstacles and dynamic convex obstacles whose boundaries do not rotate.
In general, this problem is difficult because the presence of convex obstacles makes the feasible space of trajectories nonconvex.
The performance of the algorithm is benchmarked using Monte Carlo simulations, and experimental results are presented that demonstrate the use of the method to plan collision-free multicopter trajectories in milliseconds in environments with both static and dynamic obstacles.
\end{abstract}

\section{Introduction} \label{sec:intro}

A key enabler of the use of autonomous systems in real-world situations is a fast method for generating state and input feasible trajectories between desired states.
This problem is know as the motion planning problem, and is a well researched area that includes numerous methods for the generation of such trajectories.
In particular, sampling-based methods such as rapidly exploring random trees (RRT) \cite{lavalle1998rapidly}, probabilistic roadmaps (PRM) \cite{kavraki1994probabilistic}, and fast marching trees (FMT) \cite{janson2015fast}, have been used with great success to construct collision-free paths between desired states.
Such methods are typically performed by sampling the state space of the system and attempting to connect feasible sampled nodes with simple trajectories that do not collide with obstacles.
Performing collision detection on both the sampled nodes and the trajectories that connect them is often considered the most computationally expensive step in the motion planning process, and will be the focus of this paper.

Previous work focusing on the reduction of collision detection time includes \cite{bialkowski2016efficient}, which presents an algorithm that involves using distance information from previously sampled nodes to avoid performing explicit node-obstacle collision detection when possible.
In \cite{lopez2017aggressive} this idea is adapted to reduce collision detection time for multicopter trajectories by computing overlapping collision-free spheres around the trajectory based on the maximum velocity of the vehicle and distance to the nearest obstacle at each sample point.

Rather than generating a number of multicopter trajectories and then checking each one for collisions, the authors of \cite{liu2016high} first compute a series of overlapping, obstacle-free polyhedrons and then generate a series of trajectory segments that remain inside the polyhedrons.
In \cite{chen2016online}, the authors take a similar approach by using an octree-based representation of the environment in order to enforce corridor constraints on each trajectory segment generated.
A third approach to collision avoidance for aggressive flight is explored in \cite{zhang2018p}, where a dense set of alternative trajectories to some desired trajectory are precomputed, allowing for one of the alternative trajectories to be chosen if a collision is predicted along the desired trajectory.
In this case, a collision is determined by comparing the distance of each point along the discretized candidate trajectory to the points in a point cloud generated by a laser scanner.

In contrast to methods concerned only with planning in static environments, the authors of \cite{augugliaro2012generation} leverage sequential convex programming to compute trajectories for multiple quadcopters that do not collide, allowing for dynamic formation changes. 
In \cite{nageli2017real} a method for dynamic obstacle collision avoidance is presented that models the obstacles as ellipsoids and incorporates them as nonconvex constraints in a model predictive controller.

In this paper we are interested in reducing the computational time required to find feasible trajectories for multicopters in order to enable high-speed flight in cluttered, unknown environments (e.g. when navigating a forest).
Fast trajectory generation is a requirement in these scenarios due to the limited range of onboard sensors and limited onboard computational power.
For example, obstacles can often be occluded or unexpectedly change position, requiring an immediate, agile response to avoid a collision if flying at high speeds.
Furthermore, due to the constrained onboard computational power of aerial vehicles, efficient algorithms are often required in order to achieve acceptable performance.

To this end, we propose a computationally efficient method for quickly evaluating whether a candidate trajectory collides with obstacles in the environment.
We limit ourselves to evaluating multicopter trajectories similar to those described in \cite{mueller2015computationally}, which describe the vehicle's position as a fifth order polynomial in time.
These trajectories result in the minimum average jerk over the duration of the trajectory, and are particularly useful because they can be generated and checked for input feasibility with little computation.
Unlike other collision detection methods that discretize the trajectory in time and perform a number of static collision checks at each sample point (e.g. as detailed in \cite{lavalle2006planning}), our method leverages a continuous-time representation of the trajectory to rapidly perform continuous-time collision detection.

\section{System model} \label{sec:sysModel}
We follow \cite{mueller2015computationally} in modeling the multicopter as a six degree of freedom rigid body with acceleration $\acc\in \mathbb{R}^3$ (written in the inertial coordinate frame) and orientation $\rotation$, where $\rotation$ represents the rotation matrix that rotates vectors in the body-fixed frame to the inertial frame.
Gravity is denoted $\gravity \in \mathbb{R}^3$ (written in the inertial frame), and the mass-normalized total thrust force $\thrust \in \mathbb{R}$ acts in the $\zDir$ direction, where $\zDir$ is the body-fixed thrust direction.
We assume that the angular velocity of the vehicle $\angVel$ is controlled by a high-bandwidth low-level controller such that the angular velocity converges very rapidly to a desired value, and may thus be treated as an input to the system.
The translational dynamics and attitude dynamics of the multicopter are then
\begin{equation}
\acc = \rotation \zDir \thrust + \gravity,\qquad
\rotationDot = \rotation \skewSymm{\angVel}
\end{equation}
where $\skewSymm{\angVel}$ is the skew-symmetric form of the angular velocity vector $\angVel$ so that $\skewSymm{\angVel}\bs v = \angVel \times \bs v$ for any vector $\bs v$.

Using this model, it can be shown that kinematically feasible polynomial trajectories in time can be generated using the differential flatness property of multicopter dynamics \cite{mellinger2011minimum}.
Specifically, we plan trajectories by defining the components of the jerk $\jerk(t)$ as second order polynomials in time between time $t=0$ and $t=\finalTime$.
As described in \cite{mueller2015computationally}, this results in trajectories that minimize the average jerk over the trajectory.
The thrust $\thrust$ and angular velocity $\angVel$ are then written as a function of $\acc$ and $\jerk$ as follows.
\begin{equation}
\thrust = ||\acc - \gravity||_2,\quad \begin{bmatrix}
\omega_2 \\ \omega_1 \\ 0
\end{bmatrix} = \frac{1}{\thrust}\begin{bmatrix}
1 & 0 & 0\\ 0 & 1 & 0 \\ 0 & 0 & 0
\end{bmatrix}\rotation^{-1} \jerk
\end{equation}
where $\omega_1$ and $\omega_2$ are the components of angular velocity perpendicular to the thrust direction (i.e. the roll and pitch rates).
Note that the angular velocity in the $\zDir$ direction does not affect the translational motion of the vehicle, and is taken to be $\omega_3 = 0$ for the rest of the paper.

The position and velocity of the multicopter are defined as $\pos$ and $\vel$ respectively, and are both in $\mathbb{R}^3$ and written in the inertial frame.
Let $\pos(0)$, $\vel(0)$, and $\acc(0)$ be the position, velocity, and acceleration of the vehicle at the start of the trajectory.
Because the minimum average jerk trajectory is achieved when each component of the jerk is a second order polynomial in time, the trajectories of the states of the system follow as
\begin{align}\label{eq:posParams}
\begin{bmatrix}
\pos(t) \\ \vel(t) \\ \acc(t)
\end{bmatrix} = \begin{bmatrix}
\frac{\bs \alpha}{120}t^5 + \frac{\bs \beta}{24}t^4 + \frac{\bs \gamma}{6}t^3 + \frac{\acc(0)}{2}t^2 + \vel(0)t + \pos(0) \\
\frac{\bs \alpha}{24}t^4 + \frac{\bs \beta}{6}t^3 + \frac{\bs \gamma}{2}t^2 + \acc(0)t + \vel(0) \\
\frac{\bs \alpha}{6}t^3 + \frac{\bs \beta}{2}t^2 + \bs \gamma t + \acc(0)
\end{bmatrix}
\end{align}
where $\bs \alpha$, $\bs \beta$, $\bs \gamma \in \mathbb{R}^3$ are linear functions of $\pos(\finalTime)$, $\vel(\finalTime)$, and $\acc(\finalTime)$.

A method for quickly checking whether a given trajectory satisfies bounds on the minimum and maximum thrust $\thrust$ and the magnitude of the angular velocity $\angVel$ is given in \cite{mueller2015computationally}, to which we refer the reader for further discussion.

\section{Algorithm for static obstacle collision detection}\label{sec:collision}

In this section we describe the collision detection algorithm.
All obstacles are assumed to be convex; nonconvex obstacles may be approximated by defining them as a union of convex obstacles.
In general, the presence of convex obstacles results in the feasible space being nonconvex, making the trajectory generation and collision detection problem difficult to perform using traditional optimization methods. 

We first review a method used to check whether a polynomial trajectory lies on one side of a plane, which is defined by a point $\planePos$ and unit normal $\dir$ (both written in the inertial frame).
The distance of the trajectory from the plane can be computed as
\begin{equation}
\dist(t) = \dir^T (\pos(t) - \planePos)
\end{equation}

Furthermore, the critical points of $\dist(t)$ can be computed by differentiating with respect to $t$ and finding the roots of the resulting equation:
\begin{equation}\label{eq:distDot}
\distDot(t) = \dir^T \vel(t) = c_4 t^4 + c_3 t^3 + c_2 t^2 + c_1 t + c_0
\end{equation}
where
\begin{equation}
\begin{gathered}
c_4 = \tfrac{1}{24}\dir^T \bs\alpha, \qquad c_3 = \tfrac{1}{6}\dir^T \bs\beta, \qquad c_2 =  \tfrac{1}{2}\dir^T \bs\gamma\\
\quad c_1 = \dir^T \acc(0), \qquad c_0 = \dir^T \vel(0)
\end{gathered}
\end{equation}

The trajectory $\pos(t)$ is defined only between $t=0$ and $t =\finalTime$, so the critical points of $\dist(t)$ occur between and include the start and end of the candidate trajectory. The set of critical points $\extSet$ is then defined as
\begin{equation}\label{eq:extSet}
\extSet = \{t_i : t_i \in [0, \finalTime],\; \distDot(t_i) = 0\} \cup \{0, \finalTime\}
\end{equation}

If the distance $\dist(t)$ at each critical point is positive, this indicates that $\pos(t)$ does not cross the plane.
Because $\distDot(t)$ is a fourth order polynomial in time, its roots can be found in closed form, meaning that $\extSet$ can be found with very little computation.

We now extend this method to detect collisions with convex obstacles.
The given convex obstacle $\obs$ and polynomial trajectory $\pos(t)$ are required to have the following two properties.
First, it must be possible to check whether a specific point $\pos(t_0)$ is inside $\obs$.
Second, assuming $\pos(t_0) \notin \obs$, it must be possible to define a separating plane between $\pos(t_0)$ and $\obs$ (defined with point $\planePos$ unit normal $\dir$).
Thus, if $\pos(t)$ is found to not cross the separating plane, it is guaranteed to not collide with $\obs$.

Algorithm \ref{algo:collision} leverages this property to verify whether an individual segment of a given trajectory is in collision with a given obstacle.
The algorithm begins by checking whether the end points of the trajectory $\pos(0)$ and $\pos(\finalTime)$ are inside the obstacle (lines 4-5), followed by a call to \textproc{CheckSection}, which returns whether the given section is feasible, infeasible, or whether the feasibility of the section cannot be determined (line 6).
For each call to \textproc{CheckSection}$(\sectionStartTime,\sectionEndTime)$, a time $\splitTime$ between $\sectionStartTime$ and $\sectionEndTime$ is chosen which divides the trajectory in two.
We choose $\splitTime$ to be the average of $\sectionStartTime$ and $\sectionEndTime$, as it will evenly divide the trajectory into two sub-trajectories in time (line 8).

For each section checked recursively by \textproc{CheckSection}, $\pos(\splitTime)$ is first checked for feasibility (line 9), and then the minimum resolution of the section $\minTimeSection$ is checked (line 11).
The parameter $\minTimeSection$ serves to terminate the algorithm early in order to prevent excessive time being spent checking any particular candidate trajectory, and limits the recursive depth of the algorithm.
This end condition can be reached in the case where the candidate trajectory passes sufficiently close to the obstacle, requiring the trajectory to be split into a large number of sub-trajectories to be checked.

Next, the unit normal $\dir$ and location $\planePos$ of a separating plane are found (line 13). Although there are many possible ways to find a separating plane, in our implementation we choose $\planePos$ such that it is the minimum distance point to $\pos(\splitTime)$ located in $\obs$.
The unit normal of the plane $\dir$ is then chosen such that the resulting plane lies on the obstacle boundary at $\planePos$ and points from $\planePos$ to $\pos(\splitTime)$.
The times $\extSet$ at which the critical points of the distance of the trajectory from the resulting separating plane occur are then computed by solving the corresponding fourth order polynomial given by \eqref{eq:distDot} (line 15).
Once $\extSet$ is computed, the two sections of the trajectory occurring before and after $\splitTime$ are each checked for feasibility.
Let $\extSetLow$ be the elements of $\extSet$ in $(\sectionStartTime, \splitTime)$ and $\extSetHigh$ be the elements of $\extSet$ in $(\splitTime, \sectionEndTime)$. 

\begin{figure}
	\includegraphics[width=\columnwidth]{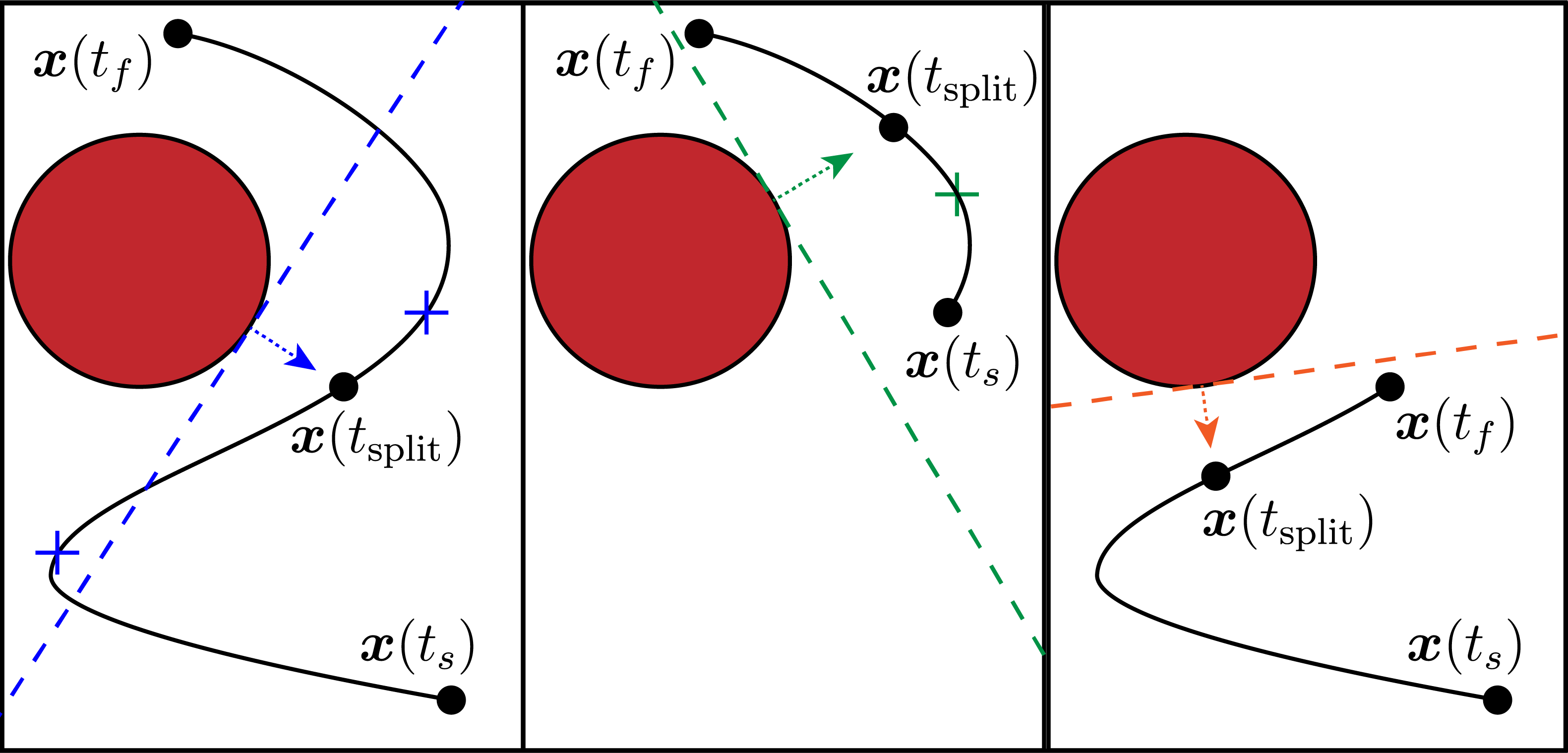}
	\centering
	\caption{A graphical depiction of Algorithm \ref{algo:collision}. Three sequential calls to \textproc{CheckSection} (as defined in Algorithm \ref{algo:collision}) are shown. The red circle represents the convex obstacle and the solid black line represents the trajectory in time. In the first call to \textproc{CheckSection} (shown in the left panel), two critical points (drawn as crosses) of the distance to the separating plane (drawn as a dashed line) are found. The trajectory is found to cross the separating plane between $\sectionEndTime$ and the critical points occurring after $\splitTime$, prompting a recursive call to \textproc{CheckSection}. As shown in the middle panel, this sub-trajectory is found to not collide with the obstacle because it lies entirely on the opposite side of the newly computed separating plane. Next, the original trajectory (left) is again found to cross the separating plane between $\sectionStartTime$ and $\splitTime$, leading to a second recursive call to \textproc{CheckSection}. As shown in the right panel, this sub-trajectory is also found to lie entirely on the opposite side of the newly computed separating plane, proving that the entire trajectory does not collide with the obstacle.}
	\label{fig:collisionCheck}
\end{figure}

The section of the trajectory between $\splitTime$ and $\sectionEndTime$ is first checked for feasibility by iterating forward in time over $\extSetHigh$ and checking whether each critical point of $\dist(t)$ lies on the feasible side of the separating plane (lines 17-18).
If a critical point is found to lie on the obstacle side of the plane, the section between the previous critical point (already determined to be on the feasible side of the plane) and $\sectionEndTime$ cannot be guaranteed to be feasible and is recursively checked with \textproc{CheckSection} (line 19).
Finally, the section of the trajectory between $\sectionStartTime$ and $\splitTime$ is checked for feasibility in a similar manner by iterating backwards in time over $\extSetLow$ (lines 26-28).
A graphical representation of Algorithm \ref{algo:collision} is shown in Figure \ref{fig:collisionCheck}.

Note that Algorithm \ref{algo:collision} treats $\pos(t)$ as the trajectory of a point.
In order to detect collisions between $\obs$ and a real multicopter, we define a sphere of radius $\quadRadius$ that contains the vehicle, and enlarge $\obs$ by $\quadRadius$ in each direction.
Additionally, because polynomials of order greater than four do not have closed form solutions except in special cases, greater computation time would be required to find the critical points of any higher order position trajectories (e.g. as used in \cite{richter2016polynomial}).

\begin{algorithm}
	\caption{Trajectory Collision Detection}
	\label{algo:collision}
	\begin{algorithmic}[1]
		\State \textbf{input:} Candidate trajectory parameters $\bs \alpha,\;\bs\beta,\;\bs\gamma$, initial conditions $\pos(0),\;\vel(0)\;\acc(0)$, minimum checking time $t_{min}$, convex obstacle $\obs$
		\State \textbf{output:} feasible, infeasible, or indeterminable
		
		\Function{CollisionCheck}{}
		\If{$\pos(0)$ or $\pos(\finalTime)$ inside obstacle}
		\State \textbf{return} infeasible
		\EndIf
		\State \textbf{return} \textproc{CheckSection}($0$, $\finalTime$)
		\EndFunction

		\Function{CheckSection}{$\sectionStartTime$, $\sectionEndTime$}
		\State $\splitTime \gets \frac{\sectionStartTime+\sectionEndTime}{2}$
		\If{$\pos(\splitTime)$ inside obstacle}
		\State \textbf{return} infeasible
		\ElsIf{$\sectionEndTime-\sectionStartTime < \minTimeSection$}
		\State \textbf{return} indeterminable
		\EndIf
		\State Find plane separating $\pos(\splitTime)$ and obstacle
		\State $\dist(t) \gets$ distance of $\pos(t)$ from separating plane
		\State $\extSetHigh \gets$ critical points of $\dist(t)$ from $\splitTime$ to $\sectionEndTime$
		\State Sort $\extSetHigh$ ascending
		\For{$t_i$ in $\extSetHigh$, skipping $\splitTime$}
		\If{$\pos(t_i)$ is on obstacle side of plane}
		\State result $\gets$ \textproc{CheckSection}($t_{i-1}$, $\sectionEndTime$)
		\If{result is feasible}
		\State \textbf{break}
		\Else
		\State \textbf{return} result
		\EndIf
		\EndIf
		\EndFor
		
		\State $\extSetLow \gets$ critical points of $\dist(t)$ from $\splitTime$ to $\sectionStartTime$
		\State Sort $\extSetLow$ descending
		\For{$t_i$ in $\extSetLow$, skipping $\splitTime$}
		\If{$\pos(t_i)$ is on obstacle side of plane}
		\State \textbf{return} \textproc{CheckSection}($\sectionStartTime$, $t_{i-1}$)
		\EndIf
		\EndFor
		\State \textbf{return} feasible
		\EndFunction
	\end{algorithmic}
\end{algorithm}

\section{Performance measures}\label{sec:perfMeas}

In this section we provide two simulations used to benchmark the performance of the proposed algorithm.\footnote{An implementation can be found at \url{https://github.com/nlbucki/RapidQuadcopterCollisionDetection}}
The algorithm was implemented in C++ and compiled with GCC version 5.4.0 with the highest speed optimization settings.
All simulations were ran as a single thread on a laptop with a 1.80GHz Intel i7-8550U processor.

\subsection{Monte Carlo simulation with random obstacles}
First, a Monte Carlo simulation was conducted in order to characterize the computational time required to perform collision detection on a single candidate trajectory.
The methods of \cite{mueller2015computationally} are used to generate the candidate trajectories and check whether the generated trajectory satisfies some given input bounds.
Any trajectory requiring a mass-normalized thrust that is not between \SI{5}{\meter\per\second\squared} and \SI{30}{\meter\per\second\squared} or that requires an angular velocity of greater than \SI{20}{\radian\per\second} is discarded.

Candidate trajectories are generated with a fixed initial position of $\pos(0) = (0,0,0)$.
The final position, initial and final velocity, and initial and final acceleration along each axis are generated from uniform distributions over the intervals (\SI{-4}{\meter}, \SI{4}{\meter}), (\SI{-4}{\meter\per\second}, \SI{4}{\meter\per\second}), and (\SI{-4}{\meter\per\second\squared}, \SI{4}{\meter\per\second\squared}) respectively.
The length of time of the trajectory is sampled uniformly at random between \SI{0.2}{\second} and \SI{4}{\second}.
A sphere with radius sampled uniformly at random on (\SI{0.1}{\meter}, \SI{1.5}{\meter}) and positions sampled uniformly at random on (\SI{-4}{\meter}, \SI{4}{\meter}) along each axis is used as an obstacle.
The minimum collision detection time per section $\minTimeSection$ is chosen to be \SI{2}{\milli\second}.

For $10^9$ such trials, the average time required to detect collisions was \SI{1.44}{\micro\second}, and of the candidate trajectories, 95.99\% did not collide with the obstacle.
Table \ref{tab:monte} shows the computation time required depending on whether the trajectory was found to be feasible, infeasible, or of indeterminable feasibility.

\begin{table}[]
	\centering
	\caption{Average collision detection time.}
	\label{tab:monte}
	\begin{tabular}{|c|c|c|c|}
		\hline
		& Feasible                                                                  & Infeasible                                                                & Indeterminable                                                            \\ \hline
		Fraction of trajectories & 95.99\%                                                                   & 4.01\%                                                                    & $<0.01$\%                                                                    \\ \hline
		Collision detection time           & \SI{1.44}{\micro\second} & \SI{1.36}{\micro\second} & \SI{11.59}{\micro\second} \\ \hline
	\end{tabular}
\end{table}

\subsection{Monte Carlo simulation with constant obstacles}
\begin{figure}
	\includegraphics[width=\columnwidth]{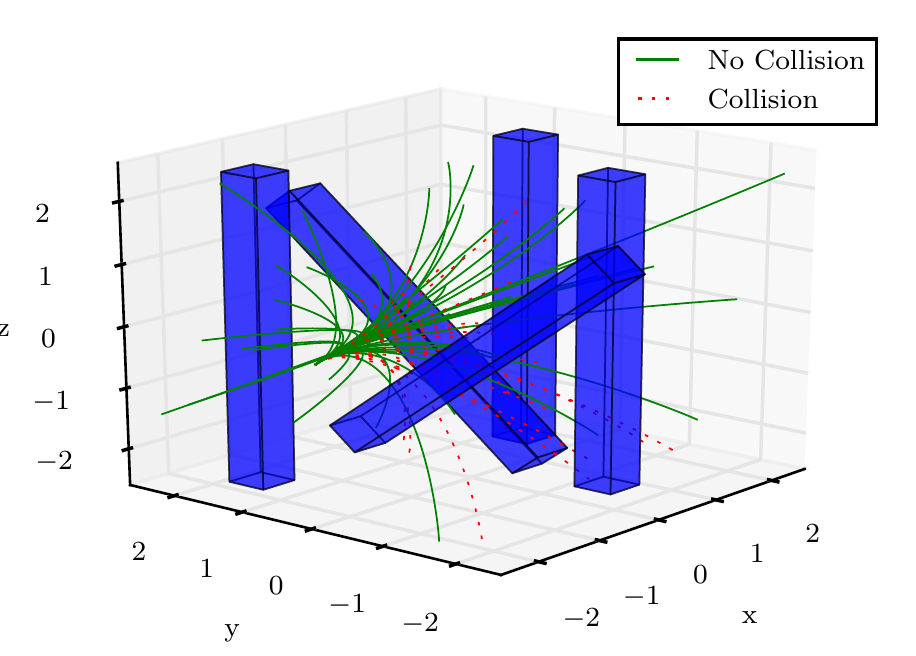}
	\centering
	\caption{Visualization of obstacle distribution and stopping trajectories. Obstacles are represented by blue rectangular prisms. Solid green lines represent the collision free trajectories from a single batch of 100 trajectories, and dotted red lines represent the trajectories that would collide with an obstacle. On average, the first feasible stopping trajectory was found in \SI{14.6}{\micro\second}.}
	\label{fig:forestPerf}
\end{figure}

A second Monte Carlo simulation involving generating collision free trajectories that bring the vehicle to rest was also conducted.
This scenario is of interest in the case where, for example, the vehicle must perform an emergency stopping maneuver (e.g. when an unexpected obstacle appears in the path of the vehicle while flying at high speed).
The simulation is run in batches of 100 candidate trajectories, where each candidate trajectory starts from the same initial state and ends at rest at a position sampled uniformly at random along each axis on (\SI{-2.5}{\meter}, \SI{2.5}{\meter}).
For each batch, the initial position of the vehicle is constrained to be (\SI{-2.5}{\meter}, \SI{0}{\meter}, \SI{0}{\meter}), the initial velocity and acceleration in the x-direction are sampled uniformly at random on (\SI{2}{\meter\per\second}, \SI{8}{\meter\per\second}) and \SI{4}{\meter\per\second\squared}, \SI{10}{\meter\per\second\squared}) respectively, and the initial velocity and acceleration in the y- and z-directions are sampled uniformly at random on (\SI{-2}{\meter\per\second}, \SI{2}{\meter\per\second}) and (\SI{-2}{\meter\per\second\squared}, \SI{2}{\meter\per\second\squared}) respectively.
The length of time of the candidate trajectories is sampled uniformly at random between \SI{0.5}{\second} and \SI{2}{\second}.
The positions and orientations of the obstacles, represented as five long rectangular prisms, are fixed as shown in Figure \ref{fig:forestPerf}, which additionally shows the candidate trajectories of a single batch.

One million batches were simulated.
The average time required to find the first collision free trajectory was \SI{14.6}{\micro\second}.
On average, each trajectory required \SI{0.1}{\micro\second} to generate, \SI{0.5}{\micro\second} to check for satisfaction of constraints on the total thrust and angular velocity, and \SI{7.7}{\micro\second} to detect any collisions with the five obstacles.
For each batch, an average of 60.2\% of generated trajectories were collision free.

\section{Dynamic Obstacle Collision Detection}\label{sec:dynAvoid}

In the previous section we showed that our algorithm is capable of detecting collisions between given quadcopter trajectories and static convex obstacles in an environment.
This method is easily extended to detect collisions with dynamic obstacles whose boundaries do not rotate, and whose position trajectories are described by fifth order or below polynomial in time.
Example applications of this method include detecting collisions between two quadcopters with different trajectories or between a projectile and a moving quadcopter.

Let $\obsTraj(t)$ be the predicted trajectory of the center of a given obstacle.
The relative position of the obstacle and the quadcopter at any time $t$ is then
\begin{equation}
	\diffTraj(t) = \pos(t) - \obsTraj(t)
\end{equation}
where each component of $\diffTraj(t)$ will a polynomial in time if each component of both $\pos(t)$ and $\obsTraj(t)$ are also polynomials in time.

Recall that we model the quadcopter as a sphere with radius $\quadRadius$.
A collision between the quadcopter and the dynamic obstacle occurs if $\diffTraj(t)$ intersects with an obstacle centered at the origin of the same size as the dynamic obstacle but enlarged by $\quadRadius$ in each direction.
Because the boundary of the obstacle is required to not rotate, the same methods described in Section \ref{sec:collision} may be used to detect collisions between $\diffTraj(t)$ and the enlarged obstacle centered at the origin.
Obstacles with boundaries that do rotate may be straight-forwardly encoded by enclosing them convex shapes with boundaries that do not rotate (e.g. spheres) at the penalty of introducing conservatism to the collision detection.
\section{Experimental results} \label{sec:experiments}

\begin{figure*}
	\centering
	\subfloat{\centering
		\includegraphics[width=\textwidth]{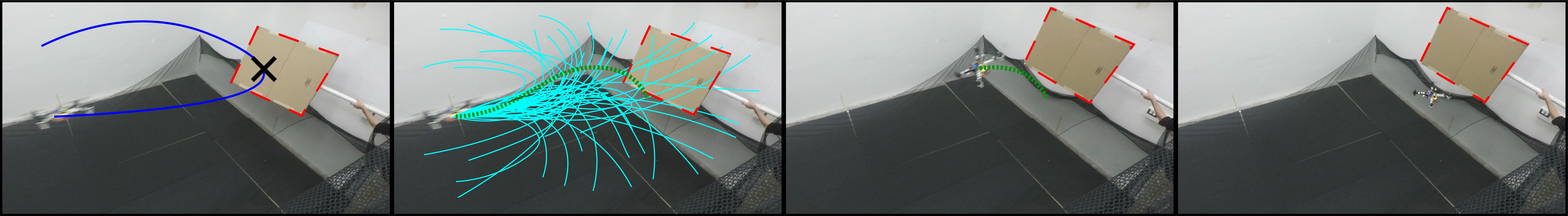}}\\[1.5px]
	\subfloat{\centering
		\includegraphics[width=\textwidth]{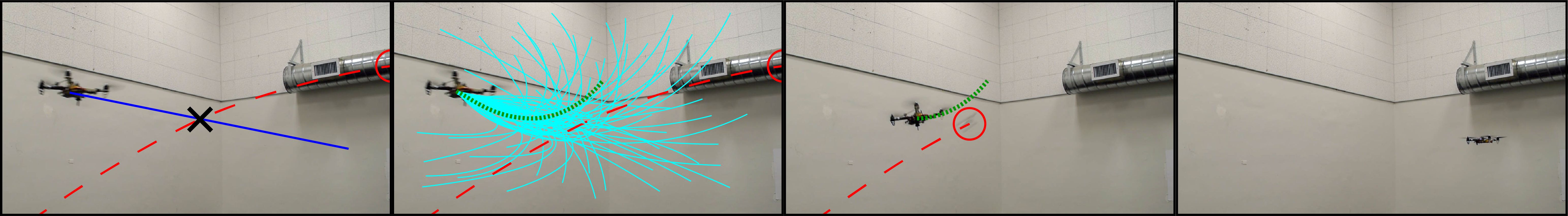}}
	\caption{A quadcopter avoiding an unexpected surface (top) and a thrown projectile (bottom). The images are from the attached video, and move forward in time from left to right. The original desired trajectory of the vehicle is shown with a solid blue line, and the position of the surface and predicted trajectory of the projectile are both shown by a dashed red line. In the first frame a collision is predicted to occur if the quadcopter remains on its current trajectory. In the second frame, a large number of alternative trajectories are generated (shown as solid cyan lines). Alternative trajectories that do not satisfy state and input constraints are discarded, and the minimum average jerk trajectory that satisfies all constraints is chosen (shown as a dotted green line). In the experiments shown, 14,610 and 2,371 candidate trajectories were generated and evaluated in \SI{15}{\milli\second} to avoid the surface and projectile respectively. In the third frame, the avoidance trajectory is tracked while continuing to detect predicted collisions with the obstacle and replanned if necessary. Finally, the fourth frame shows the vehicle successfully coming to rest without colliding with the surface (top), and generating and tracking a trajectory that brings the vehicle to the original desired end position (bottom).}
	\label{fig:imgSequence}
\end{figure*}

This section presents experimental results where the proposed algorithm is used to enable a quadcopter to avoid unexpected static and dynamic obstacles.
For the static obstacle case, we interrupt the motion of the quadcopter while following a trajectory by placing a surface in the path of the vehicle, forcing it to rapidly plan a new trajectory to avoid the collision and bring the vehicle to rest.
For the dynamic obstacle case, we throw a projectile at the vehicle while it is following a trajectory to a goal position, again forcing it to rapidly plan a new trajectory to avoid the projectile and then continue to the original goal position.
All experiments can be viewed in the attached video.

The quadcopter has a mass of \SI{685}{\gram}, and receives thrust and angular velocity commands at \SI{50}{\hertz} via radio from an offboard laptop.
Collision detection and trajectory generation is performed using the same laptop as used for benchmarking in Section \ref{sec:perfMeas}.
The position and attitude of the quadcopter and obstacles are measured using a motion capture system at \SI{200}{\hertz}.

During each controller time step, we check whether any collisions are predicted to occur between the quadcopter and the obstacle used for each experiment.
If a collision is detected, a new trajectory is generated that is not predicted to collide with the obstacle and ends at rest with zero velocity and zero acceleration.
We sample candidate end positions for the avoidance trajectory uniformly at random in a $\SI{3.2}{\meter}\times \SI{5.2}{\meter} \times \SI{1}{\meter}$ rectangular space and sample durations of the avoidance trajectory uniformly at random from \SI{0.5}{\second} to \SI{2}{\second}.
While tracking the avoidance trajectory, we continue to check for predicted collisions and generate a new avoidance trajectory if necessary.

When searching for feasible trajectories during both experiments, we generate and evaluate candidate trajectories for \SI{15}{\milli\second}, and at the end of the allocated time choose the trajectory with the minimum average jerk that satisfies all state and input constraints.
We choose the trajectory with the minimum average jerk in order to favor less aggressive trajectories.
When evaluating each candidate trajectory, we first compute the average jerk of the trajectory and reject it if it has a higher average jerk than any previously found state and input feasible trajectory.
Next, we use the methods of \cite{mueller2015computationally} to check that the total mass-normalized thrust $\thrust$ remains between \SI{5}{\meter\per\second\squared} and \SI{30}{\meter\per\second\squared} and that the maximum angular velocity remains bellow \SI{20}{\radian\per\second}, as these are the physical limits of the experimental vehicle.
We then check that the candidate trajectory stays within a box of $\SI{3.4}{\meter}\times \SI{5.4}{\meter} \times \SI{3.1}{\meter}$ to prevent the vehicle from flying into the ceiling, floor, or walls.
Finally, we check that the candidate trajectory does not collide with any obstacles using Algorithm \ref{algo:collision} with $\minTimeSection = \SI{0.002}{\second}$.

\subsection{Static obstacle avoidance}
For the static obstacle avoidance experiment, we define the static obstacle as a rectangular prism measuring $\SI{1.64}{\meter} \times \SI{1.43}{\meter} \times \SI{0.78}{\meter}$, which includes both the increase in size in each direction necessary to account for the true size of the quadcopter and a small buffer to account for trajectory tracking and estimation errors.
The quadcopter tracks two predefined trajectories that result in a roughly circular motion with an average speed of \SI{5}{\meter\per\second}, and checks these trajectories for collisions with the rectangular prism at each controller time step.
The obstacle is then moved by hand in front of the vehicle about \SI{0.5}{\second} before the vehicle would pass, causing a collision to be predicted and an avoidance trajectory to be generated that brings the vehicle to rest without colliding with the obstacle.
Figure \ref{fig:imgSequence} shows a sequence of images from the experiment.
For the experiment shown in Figure \ref{fig:imgSequence}, 14,610 candidate avoidance trajectories were evaluated in the allocated \SI{15}{\milli\second} after first predicting a collision with the obstacle (recall that the controller runs with a \SI{20}{\milli\second} period).

\subsection{Dynamic obstacle avoidance}
For the dynamic obstacle avoidance experiment, we throw a projectile at the vehicle while it is performing a rest to rest maneuver from some initial position to some final position $\finalPos$.
If a collision between the quadcopter and the projectile is predicted, the quadcopter rapidly plans a trajectory to avoid the projectile.
The projectile is thrown by hand, meaning that its trajectory can only be predicted by the system after it has been thrown.
The position $\projPos(0)$ and velocity $\projVel(0)$ of the projectile at the current controller time step are estimated using position measurements received from the motion capture system and a Kalman filter.
The position of the projectile $\projPos(t)$ is then predicted over a five second time horizon as a quadratic function of time that depends on $\projPos(0)$ and $\projVel(0)$:
\begin{equation}
\projPos(t) = \projPos(0) + \projVel(0)t + \frac{1}{2}\gravity t^2
\end{equation}

The minimum allowable distance between the center of mass of the projectile and the center of the quadcopter is chosen to be \SI{40}{\centi\meter}, which is chosen such that there is at least \SI{10}{\centi\meter} separation between the quadcopter and projectile to account for any trajectory tracking and estimation errors.
During each controller time step, we check whether the projectile is predicted to collide with the quadcopter by checking whether their relative position $\diffTraj(t)$ ever enters a sphere of radius \SI{40}{\centi\meter} centered at the origin, and begin generating an avoidance trajectories if a collision is detected.
While evaluating candidate avoidance trajectories, we not only check that the avoidance trajectory will not collide with the projectile during the maneuver, but also check that the projectile will not collide with the quadcopter after the quadcopter has reached the end position of the avoidance trajectory.

While tracking the avoidance trajectory, we try to generate sample return trajectories at each controller time step that bring the quadcopter from its current state to rest at the originally desired end position $\finalPos$.
Durations of the candidate return trajectories are sampled between \SI{0.5}{\second} and \SI{4}{\second}.
Once a feasible trajectory is found that does not collide with the projectile and ends at $\finalPos$, the avoidance trajectory is interrupted and the vehicle begins tracking the return trajectory.
Figure \ref{fig:imgSequence} shows a sequence of images from the experiment.
For the experiment shown in Figure \ref{fig:imgSequence}, 2,371 candidate avoidance trajectories were evaluated in the allocated \SI{15}{\milli\second} after first predicting a collision with the projectile.

\section{Conclusion} \label{sec:conclusion}

In this paper we presented a method for quickly detecting whether a polynomial trajectory collides with a convex obstacle.
This method can be applied to both static convex obstacles and dynamic obstacles whose boundaries do not rotate.
We used the proposed algorithm to perform rapid collision detection of multicopter trajectories, which can be modeled by fifth order polynomials in time.
The ability to rapidly assess whether a given trajectory will collide with obstacles allows for a collision-free trajectory to be found in a short period of time by generating and checking many candidate trajectories for collisions.
Because such a large number of the candidate trajectories can be generated and evaluated in such a short period of time, the vehicle is able to plan collision-free trajectories within milliseconds.
This enables the vehicle to avoid obstacles that suddenly appear while the vehicle is flying at high speeds and to avoid projectiles thrown at high speeds.

\section*{Acknowledgement}
This material is based upon work supported by the Berkeley Fellowship for Graduate Study.
The experimental testbed at the HiPeRLab is the result of contributions of many people, a full list of which can be found at \url{hiperlab.berkeley.edu/members/}.

\bibliographystyle{IEEEtran}
\bibliography{references}

\begin{thebibliography}{10}
\providecommand{\url}[1]{#1}
\csname url@samestyle\endcsname
\providecommand{\newblock}{\relax}
\providecommand{\bibinfo}[2]{#2}
\providecommand{\BIBentrySTDinterwordspacing}{\spaceskip=0pt\relax}
\providecommand{\BIBentryALTinterwordstretchfactor}{4}
\providecommand{\BIBentryALTinterwordspacing}{\spaceskip=\fontdimen2\font plus
\BIBentryALTinterwordstretchfactor\fontdimen3\font minus
  \fontdimen4\font\relax}
\providecommand{\BIBforeignlanguage}[2]{{%
\expandafter\ifx\csname l@#1\endcsname\relax
\typeout{** WARNING: IEEEtran.bst: No hyphenation pattern has been}%
\typeout{** loaded for the language `#1'. Using the pattern for}%
\typeout{** the default language instead.}%
\else
\language=\csname l@#1\endcsname
\fi
#2}}
\providecommand{\BIBdecl}{\relax}
\BIBdecl

\bibitem{lavalle1998rapidly}
S.~LaValle, ``Rapidly-exploring random trees: A new tool for path planning,''
  1998.

\bibitem{kavraki1994probabilistic}
L.~Kavraki, P.~Svestka, J.~Latombe, and M.~H. Overmars, \emph{Probabilistic
  roadmaps for path planning in high-dimensional configuration spaces}.\hskip
  1em plus 0.5em minus 0.4em\relax International Transactions on Robotics and
  Automation, 1994, vol.~12.

\bibitem{janson2015fast}
L.~Janson, E.~Schmerling, A.~Clark, and M.~Pavone, ``Fast marching tree: A fast
  marching sampling-based method for optimal motion planning in many
  dimensions,'' \emph{The International Journal of Robotics Research}, vol.~34,
  no.~7, pp. 883--921, 2015.

\bibitem{bialkowski2016efficient}
J.~Bialkowski, M.~Otte, S.~Karaman, and E.~Frazzoli, ``Efficient collision
  checking in sampling-based motion planning via safety certificates,''
  \emph{The International Journal of Robotics Research}, vol.~35, no.~7, pp.
  767--796, 2016.

\bibitem{lopez2017aggressive}
B.~T. Lopez and J.~P. How, ``Aggressive 3-{D} collision avoidance for
  high-speed navigation,'' in \emph{{IEEE} International Conference on Robotics
  and Automation (ICRA)}.\hskip 1em plus 0.5em minus 0.4em\relax IEEE, 2017,
  pp. 5759--5765.

\bibitem{liu2016high}
S.~Liu, M.~Watterson, S.~Tang, and V.~Kumar, ``High speed navigation for
  quadrotors with limited onboard sensing,'' in \emph{{IEEE} International
  Conference on Robotics and Automation (ICRA)}.\hskip 1em plus 0.5em minus
  0.4em\relax IEEE, 2016, pp. 1484--1491.

\bibitem{chen2016online}
J.~Chen, T.~Liu, and S.~Shen, ``Online generation of collision-free
  trajectories for quadrotor flight in unknown cluttered environments,'' in
  \emph{{IEEE} International Conference on Robotics and Automation
  (ICRA)}.\hskip 1em plus 0.5em minus 0.4em\relax IEEE, 2016, pp. 1476--1483.

\bibitem{zhang2018p}
J.~Zhang, R.~G. Chadha, V.~Velivela, and S.~Singh, ``P-cap: Pre-computed
  alternative paths to enable aggressive aerial maneuvers in cluttered
  environments,'' in \emph{{IEEE/RSJ} International Conference on Intelligent
  Robots and Systems (IROS)}.\hskip 1em plus 0.5em minus 0.4em\relax IEEE,
  2018, pp. 8456--8463.

\bibitem{augugliaro2012generation}
F.~Augugliaro, A.~P. Schoellig, and R.~D'Andrea, ``Generation of collision-free
  trajectories for a quadrocopter fleet: A sequential convex programming
  approach,'' in \emph{{IEEE/RSJ} International Conference on Intelligent
  Robots and Systems (IROS)}.\hskip 1em plus 0.5em minus 0.4em\relax IEEE,
  2012, pp. 1917--1922.

\bibitem{nageli2017real}
T.~N{\"a}geli, J.~Alonso-Mora, A.~Domahidi, D.~Rus, and O.~Hilliges,
  ``Real-time motion planning for aerial videography with dynamic obstacle
  avoidance and viewpoint optimization,'' \emph{IEEE Robotics and Automation
  Letters}, vol.~2, no.~3, pp. 1696--1703, 2017.

\bibitem{mueller2015computationally}
M.~W. Mueller, M.~Hehn, and R.~D'Andrea, ``A computationally efficient motion
  primitive for quadrocopter trajectory generation,'' \emph{IEEE Transactions
  on Robotics}, vol.~31, no.~6, pp. 1294--1310, 2015.

\bibitem{lavalle2006planning}
S.~LaValle, \emph{Planning algorithms}.\hskip 1em plus 0.5em minus 0.4em\relax
  Cambridge university press, 2006.

\bibitem{mellinger2011minimum}
D.~Mellinger and V.~Kumar, ``Minimum snap trajectory generation and control for
  quadrotors,'' in \emph{{IEEE} International Conference on Robotics and
  Automation (ICRA)}.\hskip 1em plus 0.5em minus 0.4em\relax IEEE, 2011, pp.
  2520--2525.

\bibitem{richter2016polynomial}
C.~Richter, A.~Bry, and N.~Roy, ``Polynomial trajectory planning for aggressive
  quadrotor flight in dense indoor environments,'' in \emph{Robotics
  Research}.\hskip 1em plus 0.5em minus 0.4em\relax Springer, 2016, pp.
  649--666.

\end{thebibliography}

\end{document}